\documentclass[runningheads]{llncs}
\usepackage{graphicx}
\usepackage{algorithm}  
\usepackage{algorithmicx}  
\usepackage{graphics}
\usepackage{multicol}
\usepackage{multirow}
\usepackage{CJKutf8}
\usepackage{subfigure}
\usepackage{booktabs}
\usepackage{amsmath} 
\usepackage{hyperref}

\makeatletter
\newcommand{\printfnsymbol}[1]{%
  \textsuperscript{\@fnsymbol{#1}}%
}
\makeatother

\begin{document}
\title{Text-guided Legal Knowledge Graph Reasoning}
%
%
 
\author{Luoqiu Li\inst{1,2}\printfnsymbol{1}
\and Zhen Bi \inst{1,2}\printfnsymbol{1}
\and Hongbin Ye \inst{1,2}\thanks{Equal contribution and shared co-first authorship.}
\and Shumin Deng \inst{1,2}\thanks{Corresponding author.}
\and Hui Chen \inst{3}
\and Huaixiao Tou\inst{3}
}
\authorrunning{L. Author et al.}
%
\institute{Zhejiang University \& AZFT Joint Lab for Knowledge Engine \and
 Hangzhou Innovation Center, Zhejiang University \and
Alibaba Group, China\\
\email{\{luoqiu.li,bi\_zhen,yehongbin,231sm\}@zju.edu.cn}\\
\email{\{weidu.ch,huaixiao.thx\}@alibaba-inc.com}
}

\maketitle              
\begin{abstract}

Recent years have witnessed the prosperity of legal artificial intelligence with the development of technologies. In this paper, we propose a novel legal application of legal provision prediction (LPP), which aims to predict the related legal provisions of affairs. We formulate this task as a challenging knowledge graph completion problem, which requires not only text understanding but also graph reasoning. To this end, we propose a novel text-guided graph reasoning approach. We collect amounts of real-world legal provision data from the Guangdong government service website and construct a  legal dataset called LegalLPP. Extensive experimental results on the dataset show that our approach achieves better performance compared with baselines.  The code and dataset are available in \url{https://github.com/zxlzr/LegalPP} for reproducibility.
\end{abstract}

\section{Introduction}

Legal Artificial Intelligence (LegalAI) mainly concentrates on applying artificial intelligence technologies to legal applications, which has become popular in recent years \cite{DBLP:conf/acl/ZhongXTZLS20}. As most of the resources in this field are presented in text forms, such as legal provisions, judgment documents, and contracts, most LegalAI tasks are based on Natural Language Processing (NLP) technologies. In this paper, we introduce a novel application of \textbf{Legal Provision Prediction (LPP)} for LegalAI.  

Legal Provision Prediction (LPP) aims to predict the related legal provisions of affairs. For example,  given an affair  ``task\_336: \begin{CJK}{UTF8}{gbsn}……超出许可业务范围或无许可证的中介服务机构发布广告的处罚\end{CJK}" (...Penalties for advertisements issued by intermediary service agencies that are beyond the scope of the licensed business or without a license), the task is to predict the most related legal provisions such as `` \begin{CJK}{UTF8}{gbsn}人才市场管理规定\_004/026/001 \end{CJK} " (Talent Market Management Regulations\_004/026/001) as the Table \ref{lpp} shows. LPP is a real-world application that plays a significant role in the legal domain, as it can reduce heavy and redundant work for legal specialists or government employees.  

Intuitively, there are many domain knowledge and concepts with well-defined rules in LegalAI, which cannot be ignored; we formulate the legal provision prediction task as a \textbf{knowledge graph completion} problem.  We regard affairs and legal provisions as entities and utilize their well-defined schema structure as relations (e.g., base\_entry\_is, base\_law\_is, etc.). In such a way, the LPP problem becomes a link prediction task in the knowledge graph (e.g., whether there exists the base\_entry\_is relation between the affair entity and the legal provision entity). Numerous link prediction approaches \cite{TransE:conf/nips/BordesUGWY13,DistMult:conf/iclr/2015,kazemi2018simple} have been proposed for knowledge graph completion; however, there are still several non-trivial challenges for LPP:

\begin{table*}[!htbp]
    \centering
     \caption{Legal Provision Prediction (LPP) task.}
    \begin{tabular}{p{2cm}|p{5cm}p{5cm}}
    \toprule
    \textbf{Type} & \textbf{Affair} & \textbf{Legal\_Provision} \\
    \midrule
   Graph Vertex (Entity) &   \begin{CJK}{UTF8}{gbsn}task\_336 \end{CJK}  &  \begin{CJK}{UTF8}{gbsn}人才市场管理规定\_004/026/001 \end{CJK}  (Talent Market Management Regulations\_004/026/001)  \\
   \midrule
   Vertex Description & \begin{CJK}{UTF8}{gbsn} 对人才中介服务机构超出许可业务范围发布广告、广告发布者为超出许可业务范围或无许可证的中介服务机构发布广告的处罚。\end{CJK} (The punishment for talent intermediary service agencies to publish advertisements beyond the scope of the licensed business, and the advertisement publishers to publish advertisements for intermediary service agencies that are beyond the scope of the licensed business or without a license.) &     
    \begin{CJK}{UTF8}{gbsn}人才中介服务机构通过各种形式、在各种媒体（含互联网）为用人单位发布人才招聘广告，不得超出许可业务范围...
    \end{CJK}  (Talent intermediary service agencies publish talent recruitment advertisements for employers in various forms and various media (including the Internet), and must not exceed the scope of the licensed business... 
    \\
    \bottomrule
    \end{tabular}
    
    \label{lpp}
\end{table*}

\begin{itemize}
\item  \emph{Text Understanding.}  Many entities in the legal knowledge graph have well-formalized description information. For example, the legal provision  "task\_336" has the description   ``\begin{CJK}{UTF8}{gbsn} ……为超出许可业务范围或无许可证的中介服务机构发布广告的处罚。\end{CJK}" (...Penalties for advertisements issued by intermediary service agencies that are beyond the scope of the licensed business or without a license). Those texts provide enriched information for understanding the affairs and legal provisions, which is quite important, and utilizing that description is of great significance. 
    
\item  \emph{Legal Reasoning.} Some complex legal provisions may require sophisticated reasoning as legal data must strictly follow the rules well-defined in law. For example, given an affair ``task\_155: \begin{CJK}{UTF8}{gbsn}市政府投资项目稽查\end{CJK}" (Audit of municipal government investment projects), human beings can quickly obtain the related legal provisions through \emph{two-hop reasoning} as ``task\_155: \begin{CJK}{UTF8}{gbsn}市政府投资项目稽查\end{CJK}" (Audit of municipal government investment projects)  is \textbf{following} ``\begin{CJK}{UTF8}{gbsn}深圳经济特区政府投资项目管理条例\end{CJK}" (Shenzhen Special Economic Zone Government Investment Project Management Article)    and  ``\begin{CJK}{UTF8}{gbsn}深圳经济特区政府投资项目管理条例\end{CJK}" (Shenzhen Special Economic Zone Government Investment Project Management Article) \textbf{has the provision of}  ``\begin{CJK}{UTF8}{gbsn}深圳经济特区政府投资项目管理条例第3节第1款\end{CJK}" (Section 1, Paragraph 3 of the Shenzhen Special Economic Zone Government Investment Project Management Regulations). 
\end{itemize}

The key to solving the issues mentioned above is combining text representation and structured knowledge with legal reasoning. To this end, we propose a \textbf{T}ext-guided \textbf{Graph} \textbf{R}easoning (\textbf{T-GraphR}) approach for this task which bridges text representation with graph reasoning. Firstly, we utilize the pre-trained language model BERT \cite{devlin2018bert} to represent entities with low dimension vectors. Then, we leverage graph neural networks (GNN) that assimilate generic message-passing inference algorithms to perform legal reasoning on the legal knowledge graph. We utilize two kinds of GNN, namely,  R-GCN \cite{schlichtkrull2018modeling} and  GAT \cite{velickovic2018graph}.  Note that our approach is a model-agnostic method and is readily pluggable into other graph neural networks approaches. 
We collect legal provisions data from Guangdong government service website\footnote{\url{https://www.gdzwfw.gov.cn/}}   and construct a dataset LegalLPP. Extensive experimental results show that our approach achieves significant improvements compared with baselines. We highlight our contributions as follows:

\begin{itemize}
\item  We propose a new legal task, namely, legal provision prediction, which requires both text representation and knowledge reasoning. 
\item  We formulate this task as a knowledge graph completion problem and introduce a novel text-guided graph reasoning approach that leveraging text and graph reasoning.
\item  Extensive experimental results demonstrate that our approach achieves better performance compared with baselines.
\item  We release the LegalLPP dataset, source code, and pre-trained models for future research purposes.
\end{itemize}

\section{Data Collection}

\subsection{Data Acquisition}
We collect all the data from the Guangdong government service website. We obtain about 140,482 raw affairs (including 1,552 unique affairs), and 4,042 laws with 269,053 legal provisions. We perform  detailed  analysis and conduct  data preprocessing procedures to address those  issues below:

\textbf{Non-standard text.} There exist a huge discrepancy for the legal provisions and affairs, including: \emph{Abbreviation}, such as ``\begin{CJK}{UTF8}{gbsn}劳动法\end{CJK}" (Labor Law) is the abbreviation of ``\begin{CJK}{UTF8}{gbsn}中华人民共和国劳动法\end{CJK}" (People's Republic of China Labor Law); \emph{Missing}, such as missing of angle quotation mark (``\begin{CJK}{UTF8}{gbsn}《》\end{CJK}" ) or the format of the version number (for example, the suffix of the ``\begin{CJK}{UTF8}{gbsn}《广东省民用建筑节能条例》（2014年修正本）\end{CJK}" ("Regulations on Energy Conservation of Civil Buildings in Guangdong Province" (Amended in 2014)). Those challenges make it difficult to establish the association between affairs and legal provisions. To handle those non-standard texts, we manually build a legal provision dictionary to normalize those non-standard texts. 

\textbf{Similar affairs}. From the raw data, we observe that a huge portion of affairs is very similar (with the same affair vertexes). Statistically, we find that the ratio of unique items to the total number of items is roughly 1:100.  We analyze those similar affairs and find that most parts of them have the same content, while only the \emph{time} in the affair is different. We merge those similar affairs in the prepossessing procedure. 

\textbf{No legal provisions.} We observe that several affairs do not have any linking legal provisions, which implies no legal provisions. This problem is mainly due to outdated laws. As the legal provision will change over time,  several old provisions may be deleted, making several affairs impossible to link.  Also, there exists a little legal provision that does not have standard formats (in general, the standard format of the legal provision is XX law, chapter X, article X, paragraph X); thus, affairs cannot be linked to those legal provisions either. We filter out those no legal provision affairs in the prepossessing procedure. 

\begin{figure*} [!htbp]
\centering
\includegraphics[width=0.7\textwidth]{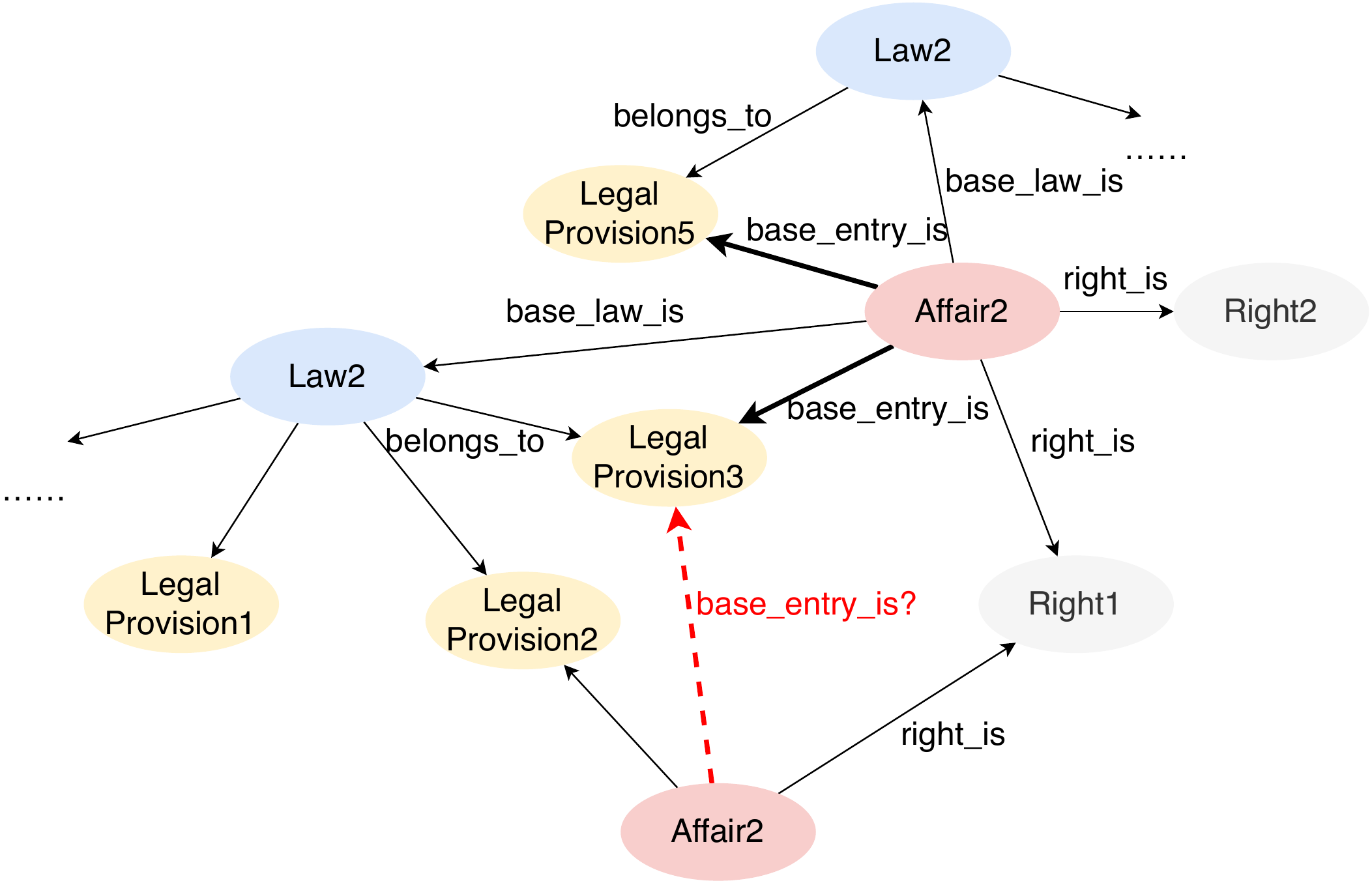}
\caption{Legal Provision Prediction as Link Prediction on Legal Knowledge Graph. Best view in color.}
\label{kg}
\end{figure*}

\begin{table}[!htbp]
    \centering
         \caption{Statistic of the legal knowledge graph. \textbf{base\_entry\_is}  is the target relation.}
    \begin{tabular}{c|cc}
    \toprule
    \textbf{Relation} & \textbf{Number} & \textbf{Description} \\
    \midrule
  \textbf{base\_entry\_is}  & 4,526& The legal provision  is related to the affair   \\
   right\_is&  1,090  &  The affair has the right   \\
   base\_law\_is& 2,152  & The law is related to the affair  \\
   belongs\_to& 182,624&  The legal provision belongs to the law   \\
  \bottomrule
    \end{tabular}

    \label{schema}
\end{table}

\subsection{Legal Knowledge Graph}
Our proposed legal task is a real-world application. As the fast updating of affairs and legal provisions, newly added affairs cannot be linked with existing legal provisions. We notice that there exist relations between affairs and laws following a well-defined schema.  From two-hop reasoning on the legal knowledge graph, it is possible to judge whether there exists a relation between an affair and legal provisions.  
In this paper, we formulate the legal provision prediction task as the link prediction problem in the legal knowledge graph. We model the \emph{legal provision}, \emph{affair}, \emph{law}, \emph{right} as entities, as shown in Figure \ref{kg}.  We detail the statistic of the legal knowledge graph in Table \ref{schema}. 

\subsection{Dataset Construction}
We randomly divide the triples with four type relations into the train, valid and test set with a ratio of 8:1:1, and we filter the triples with base\_entry\_is relation from the test set as the target test set.  The detailed number of entities, relations, and triples of the LegalLPP dataset are shown  in Table \ref{tab:dataset}.

 \begin{table}[!htbp]
    \centering
         \caption{Summary statistics of LegalLPP dataset.}
    \begin{tabular}{c|ccc}
    \toprule
    \textbf{Dataset} & \textbf{\#Rel} & \textbf{\#Ent} & \textbf{\#Triple}  \\
    \midrule
     Train(all) & 4& 151,746  & 152,307  \\
     Dev(all) & 4& 22,086 & 19,037   \\
     Test(all) & 4& 22,070 & 19,042   \\
     Test(target) & 1& 768 &  454 \\
    \bottomrule
    \end{tabular}

    \label{tab:dataset}
\end{table}

\section{Methodology}
\subsection{Problem Definition}
A knowledge graph $G$ is a set of triplets in the form $(h, r, t)$, $h, t \in \mathcal{E}$ and $r \in \mathcal{R}$ where $\mathcal{E}$ is the entity vocabulary and $\mathcal{R}$ is a collection of pre-defined relations as shown in Table \ref{schema}. We are aimed at \emph{predicting whether there exists the relation base\_entry\_is between affair entities and legal provision entities}. We construct positive triples with ground truth instances and negative triples with corrupted instances following \cite{TransE:conf/nips/BordesUGWY13}.

\subsection{Framework Overview}
\begin{figure*} \centering
\includegraphics[width=1\textwidth]{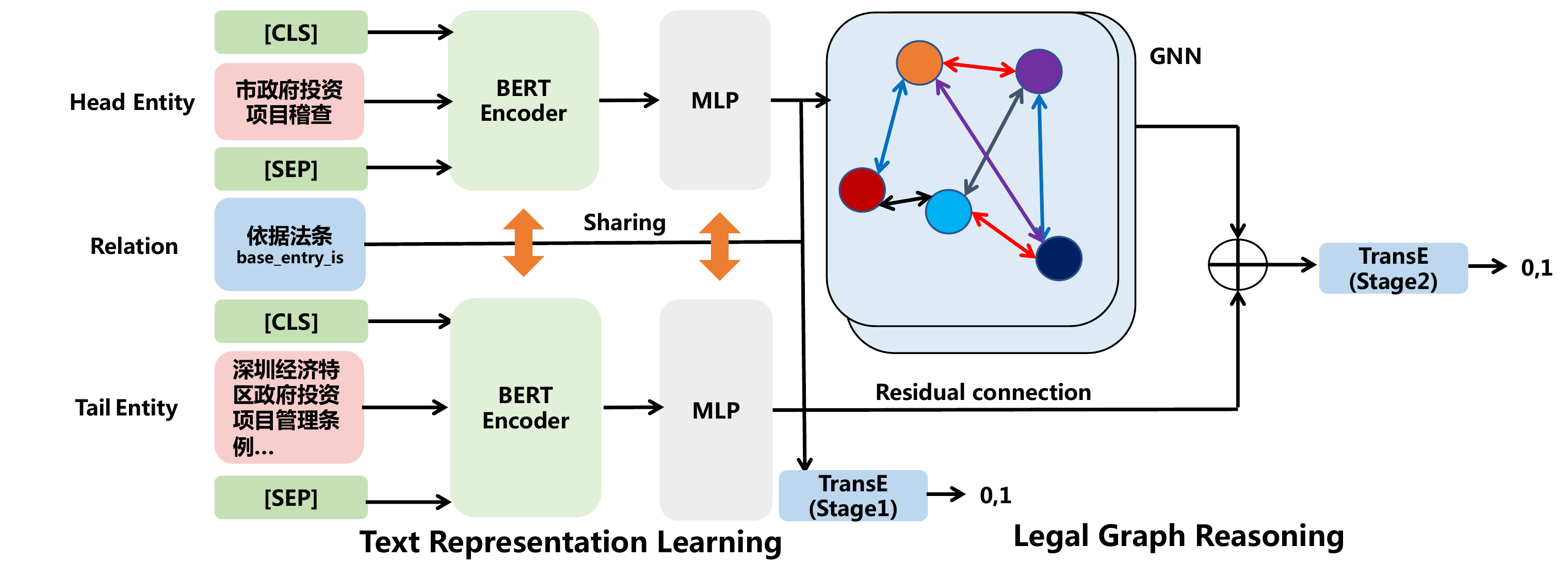}
\caption{Our approach of \textbf{T}ext-guided \textbf{Graph} \textbf{R}easoning (\textbf{T-GraphR}). TransE (Stage1) and TransE (Stage2) refers to the triple score function in text representation learning (\S \ref{text_sec}) and legal graph reasoning (\S \ref{graph_sec}), respectively. Best view in color.}
\label{arc}
\end{figure*}

Our text-guided graph reasoning approach consists of two main components, as shown in  Figure \ref{arc}. Our approach is not end-to-end as we firstly fine-tune the text representation and then leverage this feature to perform legal graph reasoning.
 
\textbf{Text Representation Learning} (\S \ref{text_sec}).  Given an affair and legal provision, we employ neural networks to encode the instance semantics into a vector. In this study, we implement the instance encoder with BERT \cite{devlin2018bert}. We then apply an MLP layer to reduce the dimension of features (the dimension of BERT-base is 768, which is not convenient for training GNN) to obtain the text representations, which is more efficient for training and inference.  We learn the text representation via fine-tuning with triple scores following TransE \cite{TransE:conf/nips/BordesUGWY13}.

\textbf{Legal Graph Reasoning} (\S \ref{graph_sec}). After obtaining the learned text representations, we employ  GNN to learn explicit relational knowledge. By assimilating generic message-passing inference algorithms with the neural-network counterpart, we can learn vertex embeddings with legal reasoning. Then we utilize a residual connection from the text representation to obtain the final representation. Finally, we utilize TransE \cite{TransE:conf/nips/BordesUGWY13}, DistMult \cite{DistMult:conf/iclr/2015} and SimplE \cite{kazemi2018simple} as triple score functions. 

\subsection{Text Representation Learning\label{text_sec}}
Given an input affair text $h$ and  legal provision text $t$, we utilize BERT \cite{devlin2018bert} to obtain the text representations as follows:
\begin{equation}
    m_{h}=\operatorname{BERT}(h),    
          m_{t}=\operatorname{BERT}(t)
\end{equation}

where $h$,$t$ are raw input text and $m_h$,$m_t$ are the output  [CLS] embeddings of BERT. We then leverage an MLP layer to reduce dimension as follows:
\begin{equation}
    v_{h} = ReLU(W^h * m_{h} + b^h),
       \qquad v_{t} = ReLU(W^t * m_{t} + b^t)
           \label{mlp}
\end{equation}

where $v_{h}$, $w_t$ are the final text representations which will then be fed into the GNN. To obtain more representative features, we finetune the text representation with TransE triple score function as:
\begin{equation}
    \operatorname{score}(h, r, t)_{transe}= \|v_{h} + v_{r} - v_{t}\|_p
    \label{transe}
\end{equation}

where $\operatorname{score}(h, r, t)$ is the score of triple $<h,r,t>$, $v_h$, $v_t$, are entity representations from Eq. \ref{mlp}, $v_r$ is \textbf{random initialized vectors}. We further analyze the empirical performance of
other triple score function of  DistMult and SimplE.  The DistMult and SimplE score function are  calculated as: 
 
\begin{equation}
    \operatorname{score}(h, r, t)_{distmult}=\sum\left(v_{h} * v_{r} * v_{t}\right) 
\end{equation}
\begin{equation}
    \operatorname{score}(h, r, t)_{simple}=\frac{\sum\left(v_{h} * v_{r} * v_{t}\right)+\sum\left(v_{h} * v_{r_{i n v}} * v_{t}\right)}{2}
    \label{simple}
\end{equation}
DistMult is a simplified version of RESCAL \cite{RESCAL:conf/icml/NickelTK11} by using a diagonal matrix to encode relation. Different from DistMult, SimplE can handle asymmetric relations. 

\subsection{Legal Graph Reasoning\label{graph_sec}}
We feed the vertex representation $v_i$ into a graph encoder to obtain the hidden vectors, which explicitly models the graph structure of the legal knowledge graph.  We use an implementation of the GNN model following GAT \cite{velickovic2018graph} and R-GCN \cite{schlichtkrull2018modeling}.  

\textbf{GAT.} The GAT model uses multiple graph attention layer connections for encoding vertexs. Specifically, GAT calculates the attention weights of neighboring nodes for aggregation. The attention weights of node $i$ and node $j$ are calculated as follows:

\begin{equation}
\alpha_{i j}=\frac{\exp \left(\overrightarrow{\mathbf{a}}^{T}\left[\mathbf{W} v_{i} \| \mathbf{W} v_{j}\right]\right)}{\sum_{k \in \mathcal{N}_{i}} \exp \left(\overrightarrow{\mathbf{a}}^{T}\left[\mathbf{W}  v_{i} \| \mathbf{W} v_{k}\right]\right)}
\end{equation}
where $. T$ represents transposition and $\|$ is the concatenation operation. Once obtained, the normalized attention coefficients are used to compute a linear combination of the features corresponding to them, to serve as the final output features for every node (after potentially applying a nonlinearity, $\sigma$):
\begin{equation}
     v_{i}^{\prime}=\sigma\left(\sum_{j \in \mathcal{N}_{i}} \alpha_{i j} \mathbf{W} v_{j}\right)
\end{equation}
where $\vec{v}_{i}^{\prime}$ is the final graph node representation. 

\textbf{R-GCN.} R-GCN utilizes multi-layer relational graph convolutional layer to represent node. The forward-pass update of an node denoted by $v_i$ in a relational  multi-graph is shown as follows:

\begin{equation}
    v_{i}^{\prime}=\sigma\left(\Sigma_{r \in R} \Sigma_{m \in N_{i}^{r}} \frac{1}{c_{i, r}} W_{r} v_{i}+W_{0} v_{i}\right)
\end{equation}
where $\vec{v}_{i}^{\prime}$ is the final graph node representation, $\mathcal{N}_{i}^{r}$ denotes the set of neighbor indices of node $i$ under relation $r \in \mathcal{R}$. $c_{i,r}$ is a problem-specific normalization constant that can either be learned or chosen in advance
(such as $c_{i, r}=\mid \mathcal{N}_{i}^{r}$). Note that, R-GCN  considers the relation of triples in the convolution process and is able to learn different aggregation weights according to different relations.

Afterwards, we add a residual connection from the output of MLP layer to the graph node representation, denoted by:
\begin{equation}
    v_i = v_i + GNN(v_i)
\end{equation}
where $v_i$ is the final entity representation of entities  leveraging both text and graph reasoning. 

Finally, we utilize the same score function of TransE, DistMult and SimpLE in the \S \ref{text_sec} to calculate triple scores. Note that, in the graph reasoning stage, $v_{h}$ and $v_t$ are combined with  both text and graph features while $v_r$ is initialized  from \textbf{the  tuned embedding in  text representation learning} (Eq. \ref{transe} and Eq. \ref{simple}). 
Though our approach is not end-to-end, the  entity embeddings (e.g., legal provisions, affairs) can be pre-computed, which is quite efficient in for inference.

\section{Experiments}
\subsection{Settings}
We conduct experiments on the LegalLPP dataset.  We use Pytorch \cite{paszke2019pytorch} to implement baselines and our approach on single Nvidia 1080Ti GPU.  We leverage Graph Deep Library\footnote{\url{https://www.dgl.ai/}} to implement all the GNN components. We utilize \emph{bert-base-Chinese}\footnote{\url{https://github.com/google-research/bert}} to represent text. We employ Adam~\cite{Kingma2015AdamAM} as the optimizer. In the text representation learning stage, the learning rate is 5e-5 with the warm-up proportion being 0.1; the batch size is 64,  the maximum sequence length of each entity's text is 128.  After 6 epochs of training, we generate 400-dimension text representations. In the legal graph reasoning stage, we set the learning rate of GNN to be 0.01. We train 4,000 epochs for GAT and R-GCN. We use  TransE as the default triple score function. 
We evaluate the performance with Mean Rank (MR), Mean Reciprocal Rank (MRR), and HIT@N (N=1,3,10).  

\subsection{Baselines}
We compare our approach with different kinds of baselines, as shown below:

\textbf{No reasoning.} We conduct TransE \cite{TransE:conf/nips/BordesUGWY13} as an   baseline. We also utilize two separate  BERT encoders to represent the text with the TransE triple score function as a baseline. 

\textbf{Graph only.} We build the legal knowledge graph and leverage GNN approaches R-GCN and GAT without text features. The graph node representation is initialized randomly. 

\subsection{Evaluation Results\label{exp}}

\begin{table*}[!htbp]
\centering
\caption{Main results on LegalLPP dataset.}
\begin{tabular}{cc|ccccc}
\toprule
 \multicolumn{2}{c|}{\textbf{Model}}&  \textbf{MR} & \textbf{MRR} & \textbf{HIT@1} & \textbf{HIT@3} & \textbf{HIT@10} \\
 \midrule
\multirow{2}{*}{No reasoning}
 & TransE & 21615.832 & 0.179 & 0.121 & 0.196 & 0.258  \\
  & BERT & \textbf{404.308} & 0.103 & 0.051 & 0.095 & 0.207 \\
\midrule
\multirow{2}{*}{Graph only} & GAT & 14790.835 & 0.187 & \textbf{0.137} & 0.209 & 0.262 \\
& R-GCN & 35767.694 & 0.175 & 0.119 & 0.187 & 0.267 \\
\midrule
\multirow{4}{*}{T-GraphR}
& GAT (TransE) & 21339.555 & \textbf{0.197} & 0.133 & \textbf{0.214} & \textbf{0.291} \\

 & GAT  (DistMult)  & 19546.152 & 0.047 & 0.011 & 0.041 & 0.119 \\
  & GAT (SimplE) & 18164.057 & 0.094 & 0.062 & 0.099 & 0.145 \\
  & R-GCN  (TransE) & 1414.584 & 0.179 & 0.126 & 0.192 & 0.242 \\
\bottomrule
\end{tabular}

\label{main}
\end{table*}

From Table \ref{main}, we observe: 

1) Our approach T-GraphR with GAT achieves the best performance. We argue that our target task is to predict the base\_entry\_is relation between affairs and legal provisions, and there are only four relations in the graph; thus, GAT, which implicitly specifying different weights to different nodes in a neighborhood can obtain better performance. 



2) Graph only approach achieves better performance than no reasoning methods BERT and TransE, 
which indicates that graph reasoning plays a vital role in legal provision prediction. 

3) Our T-GraphR approach achieves the best performance and even obtain 12.8\% hit@10 improvements compared with the text-only no reasoning model TransE.

4) The overall performance is still far from satisfactory (less than 0.3 with hit@10), and there is more room for future works.

We conduct experiments with different triple score function and report results in Table \ref{main}. We observe that TransE obtains better performance than DistMult and SimplE. We argue that the TransE model represents relations as translations, which aims to model the \textbf{inversion} and \textbf{composition} patterns;  the DistMult utilizes the three-way interactions between head entities, relations, and tail entities which aims to model the \textbf{symmetry} pattern, the SimplE model the \textbf{asymmetric} relations by considering two (head and tail) vectors only. In our LPP task, those inversion and composition patterns are common in the legal knowledge graph; thus, such translation assumption is advantageous.
\begin{table*}[!htbp]
    \centering
        \caption{Case Studies.}
    \begin{tabular}{c|p{3.5cm}p{3.8cm}p{3.8cm}}
    \toprule
    \textbf{Model} & \textbf{Affair} & \textbf{T-GraphR} & \textbf{BERT} \\
    \midrule

    Instance1&  \begin{CJK}{UTF8}{gbsn} 对占用道路、广场从事经营性车辆清洗活动的处罚\end{CJK} (Penalties for occupation of roads and plazas for cleaning vehicles) &     
    \begin{CJK}{UTF8}{gbsn}肇庆市城区市容和环境卫生管理条例\_005/053/001\end{CJK} (Regulations of Zhaoqing City on City Appearance and Environmental Sanitation\_005/053/001) &    \begin{CJK}{UTF8}{gbsn}中华人民共和国河道管理条例\_003/035/001\end{CJK} (River Regulations of the People's Republic of China\_003/035/001)\\
    \midrule
      Instance2&  \begin{CJK}{UTF8}{gbsn}  对未按规定缴纳城市生活垃圾处理费的行政处罚\end{CJK}  (Administrative penalties for failure to pay municipal solid waste disposal fees) &     
    \begin{CJK}{UTF8}{gbsn} 广东省城乡生活垃圾处理条例\_004/037/001 \end{CJK}  (Guangdong Province Urban and Rural Domestic Waste Treatment Regulations\_004/037/001) &    \begin{CJK}{UTF8}{gbsn}广东省环境保护条例\_004/056/001\end{CJK} (Guangdong Environmental Protection Regulations\_004/056/001)\\
    \bottomrule
    \end{tabular}
 
    \label{case}
\end{table*}

\subsection{Case Studies}
We present some predicted instances obtained by our model to demonstrate the generalization ability in Table~\ref{case}. Our method can predict correct legal provisions with complex surface contexts. Moreover, by reasoning on the legal knowledge graph, we can leverage the well-defined structure, which boosts performance.  However, vanilla BERT only considers text, neglecting the structured knowledge in the legal knowledge graph, which results in unsatisfactory performance.

\subsection{Entity Visualization}
\begin{figure*}[h]
\centering

\subfigure[GAT] { 
  \includegraphics[scale=0.4]{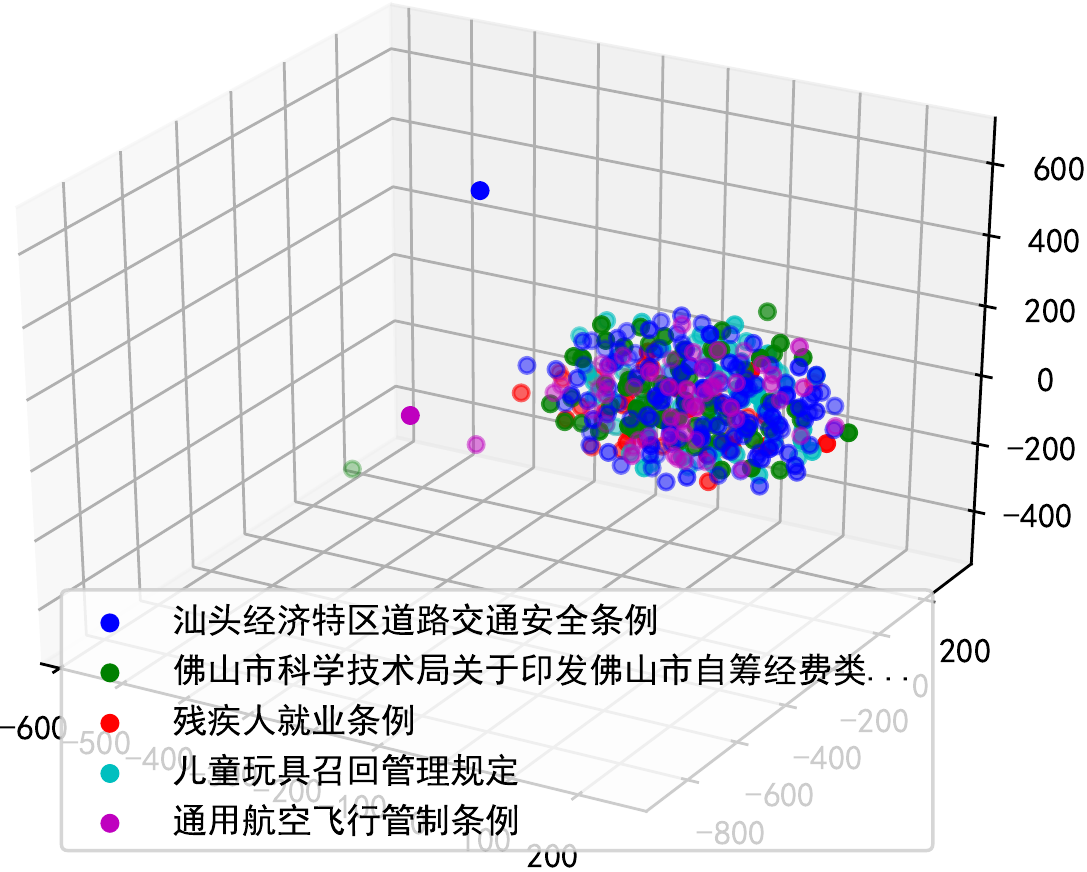}
}
\subfigure[R-GCN] { 
 \includegraphics[scale=0.4]{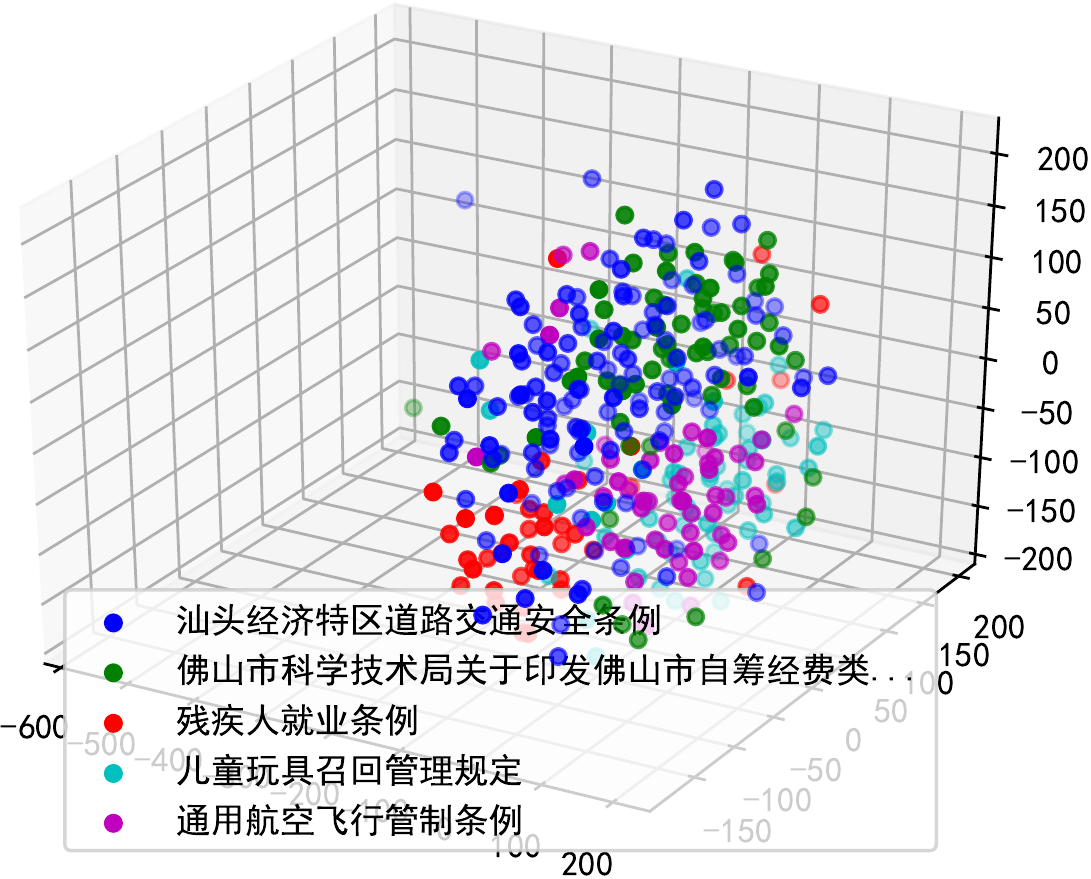}
}
\subfigure[T-GraphR (GAT)] { 
  \includegraphics[scale=0.4]{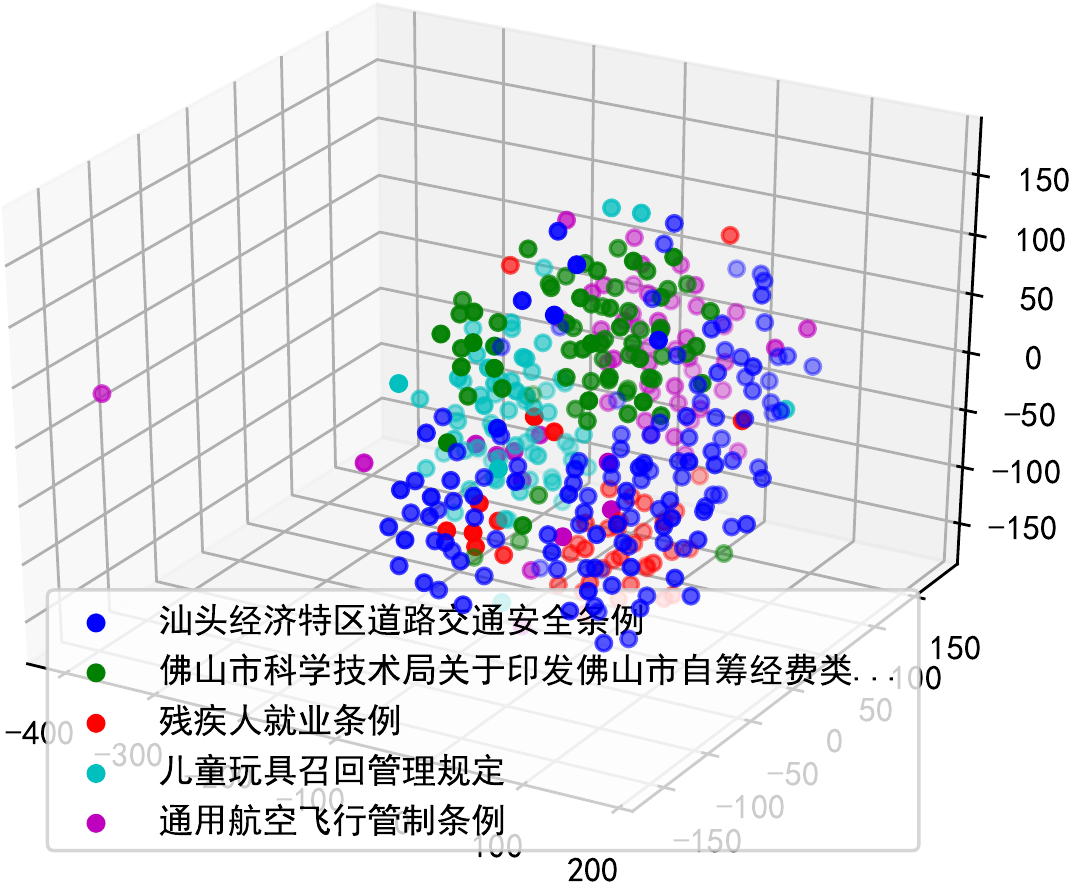}
}
\subfigure[T-GraphR (R-GCN)] { 
 \includegraphics[scale=0.4]{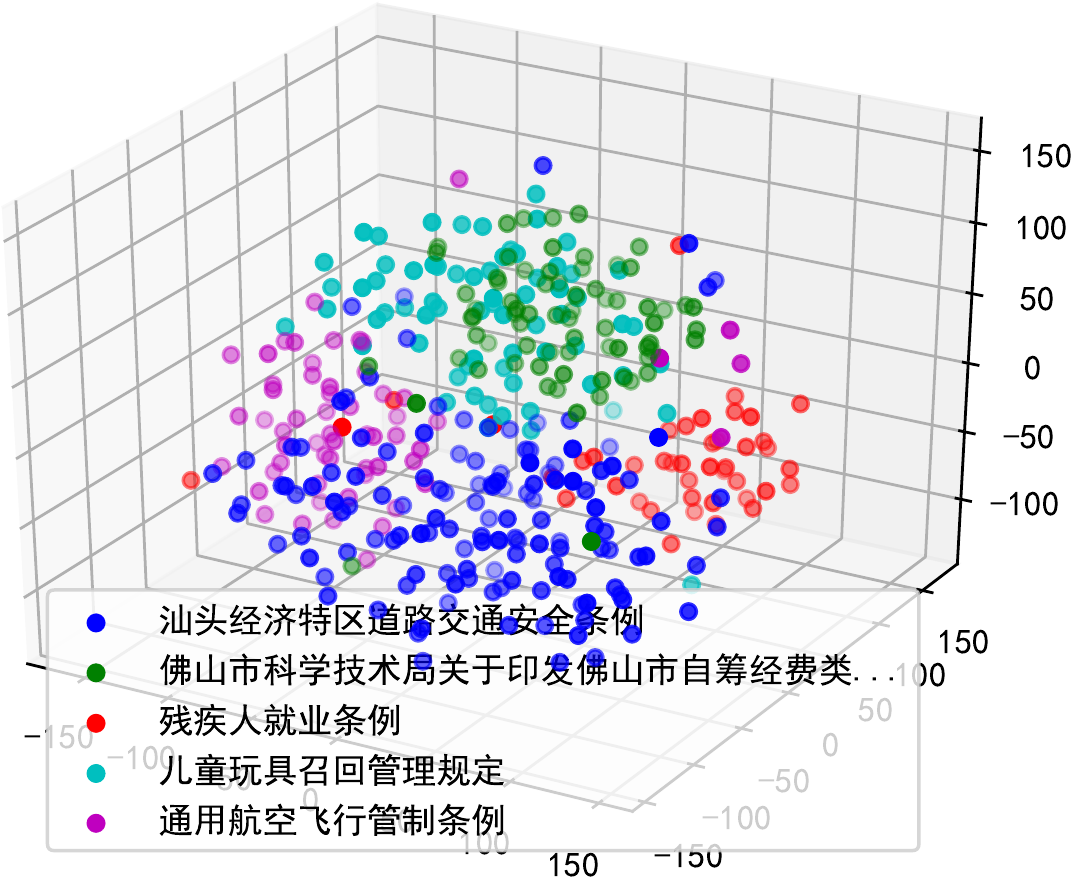}
}

\caption{T-SNE visualizations of entity (legal provision) embeddings. Figure (a) and (b)  refer to the entity embeddings of  \textbf{graph only} method, Figure (c) and (d) refer to our \textbf{T-GraphR} approach.
}
\label{vis}
\end{figure*}

To further analyze the behavior of entity representations, we utilize T-SNE \cite{maaten2008visualizing} to visualize five randomly selected entity embeddings. From Figure \ref{vis}, we find that entity embeddings of the graph only approaches have a compact data distribution, while with pre-trained LMs, entities of different types are \textbf{scattered}.    To conclude, text features can enhance the vertex's discriminative ability to enhance node representations. 

\section{Related Work}
\textbf{Knowledge Graph Completion.}
In this paper, we formulate the LPP problem as a knowledge graph completion task by link prediction. A variety of  approaches such as TransE \cite{TransE:conf/nips/BordesUGWY13}, ConvE \cite{ConvE:conf/aaai/DettmersMS018}, Analogy \cite{ANALOGY:conf/icml/LiuWY17}, RotatE \cite{RotatE:Sun2019RotatE} have been proposed to encode entities and relations into a continuous low-dimensional space \cite{zhang2020relation}. TransE \cite{TransE:conf/nips/BordesUGWY13}  regards the relation $r$ in the given fact $(h, r, t)$ as a translation from $h$ to $t$ within the low-dimensional space. RESCAL \cite{RESCAL:conf/icml/NickelTK11}  studies on matrix factorization based knowledge graph embedding models using a bilinear form as score function.  DistMult \cite{DistMult:conf/iclr/2015} simplifies RESCAL by using a diagonal matrix to encode relation.
\cite{kazemi2018simple} propose a simple tensor factorization model called SimplE through a slight modification of the Polyadic Decomposition model \cite{hitchcock1927expression}. Since the relation of the legal knowledge graph is quite small, we utilize TransE \cite{TransE:conf/nips/BordesUGWY13}, DistMult \cite{DistMult:conf/iclr/2015} and SimplE \cite{kazemi2018simple} as score functions of knowledge graph completion for computation efficiency.

\textbf{Graph Neural Networks.} 
Recently, graph neural network (GNN) models have increasingly attracted attention, which is beneficial for graph data modeling and reasoning. Some existing literature such as   R-GCN \cite{schlichtkrull2018modeling},  GAT \cite{velickovic2018graph} use GNN for structure learning. 
\cite{schlichtkrull2018modeling} introduces a relational graph convolutional networks (R-GCN) for knowledge base completion tasks that can deal with the highly multi-relational data.
\cite{velickovic2018graph}  propose a graph attention networks (GAT) that leveraging masked self-attentional layers based on neural graph networks.
However, as legal provision also has lots of text information which cannot be ignored; thus, we leverage pre-trained text representation as guidance for graph reasoning.

\section{Conclusion}
In this paper, we introduce an application of legal provision prediction, which requires text understanding and knowledge reasoning. This task can reduce heavy and redundant work for legal specialists or government employees. We formulate this task as a knowledge graph completion task and propose a text-guided graph reasoning approach. Experimental results demonstrate the efficacy of our approach, however, the task is still far from satisfactory.  

%
%
%
 \bibliographystyle{splncs04}
 \bibliography{acl2021}

\end{document}